\pdfoutput=1

\documentclass[11pt]{article}

\usepackage[final]{acl}

\usepackage{times}
\usepackage{latexsym}

\usepackage[T1]{fontenc}

\usepackage[utf8]{inputenc}

\usepackage{amsmath}
\usepackage{microtype}
\usepackage{tabularx}
\usepackage{graphicx}
\usepackage{float}
\usepackage{makecell}
\usepackage{booktabs}
\usepackage{color}
\usepackage{xargs}
\usepackage[colorinlistoftodos,prependcaption,textsize=tiny]{todonotes}

\title{Reducing Target Group Bias in Hate Speech Detectors}

 \author{Darsh J Shah ~~
Sinong Wang ~~
Han Fang ~~
Hao Ma ~~ Luke Zettlemoyer\\
{Meta AI Research}\\ 
{\{darshs,sinongwang,hanfang,haom,lsz\}}@fb.com}

\begin{document}
\newcommandx{\sinong}[2][1=]{\todo[linecolor=red,backgroundcolor=red!25,bordercolor=red,#1]{#2}}
\maketitle
\begin{abstract}
The ubiquity of offensive and hateful content on online fora necessitates the need for automatic solutions that detect such content competently across target groups. In this paper we show that text classification models trained on large publicly available datasets despite having a high overall performance, may significantly under-perform on several protected groups. On the \citet{vidgen2020learning} dataset, we find the accuracy to be 37\% lower on an under annotated Black Women target group and 12\% lower on Immigrants, where hate speech involves a distinct style. To address this, we propose to perform token-level hate sense disambiguation, and utilize tokens' hate sense representations for detection, modeling more general signals. On two publicly available datasets, we observe that the variance in model accuracy across target groups drops by at least 30\%, improving the average target group performance by 4\% and worst case performance by 13\%.
\end{abstract}

\section{Introduction}
\begin{figure}[!t]
\includegraphics[width=1.0\columnwidth]{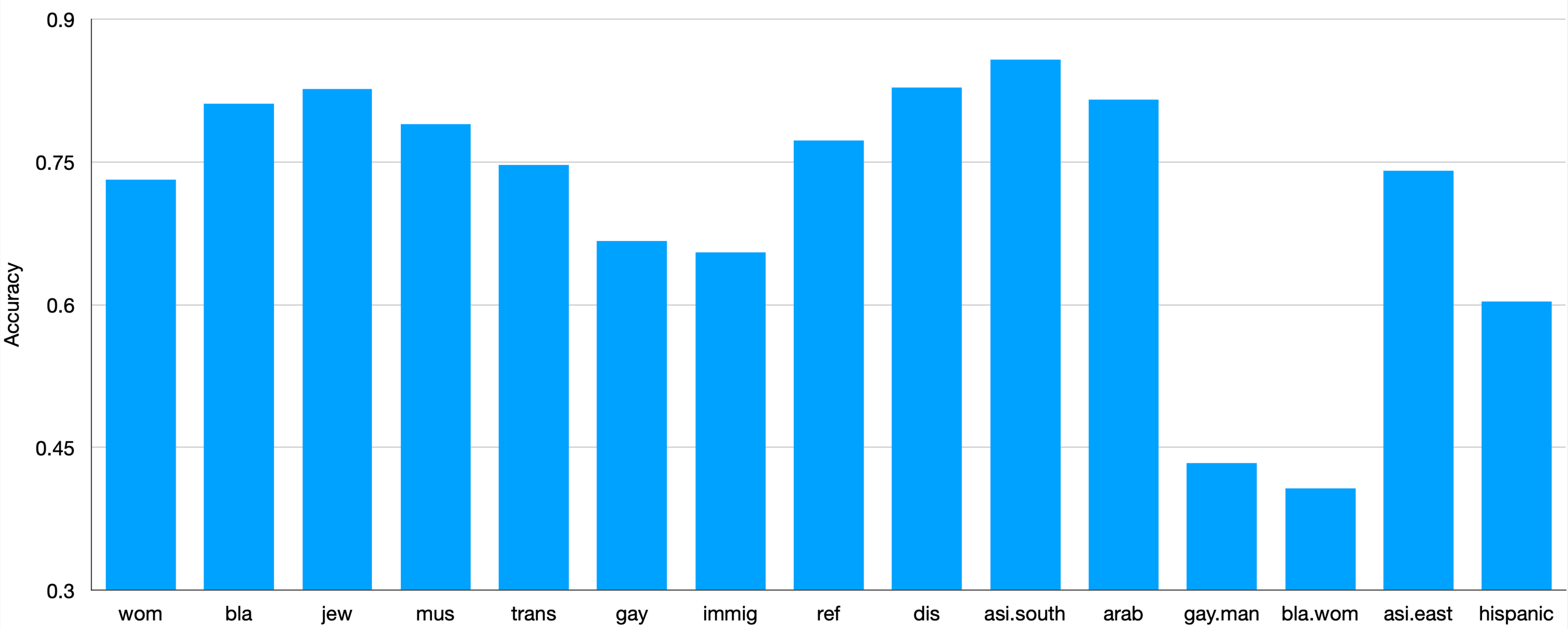}
\caption{The performance of a state-of-the-art model on hate speech detection across different target groups on \citet{vidgen2020learning}.}
\label{fig:intro}
\end{figure}

The diverse nature of hate speech against distinct target groups makes its automatic detection very challenging. In this paper, we study the impact of training machine learning models on two public hate speech datasets, where the content is organically driven by forum users, making the subsequent corpora unbalanced. While datasets should reflect content produced in the real world, we find models trained on such unbalanced datasets to perform with varying competence across target groups -- demographic segmentations, often being poorer for protected groups. For example a BERT model \citep{devlin-etal-2019-bert} trained and evaluated on the dataset in \citet{vidgen2020learning}, has a high variance in detection accuracy across different target groups, significantly under performing on attacks against Gay Men and Black Women (see Figure \ref{fig:intro}).

\begin{figure}[!h]
\includegraphics[width=1.0\columnwidth]{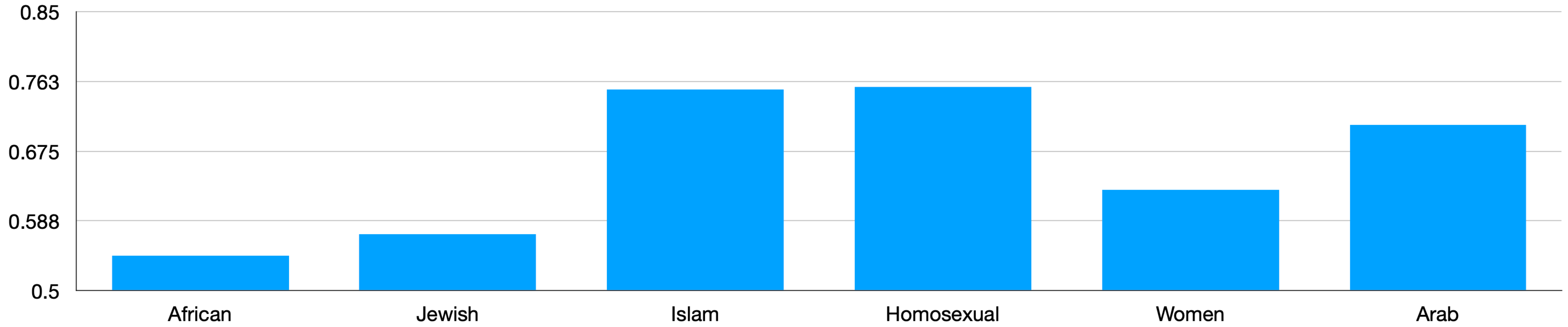}
\caption{The performance of a state-of-the-art model on hate speech detection across different target groups on \citet{mathew2020hatexplain}.}
\label{fig:intro2}
\end{figure}

Our analysis of this bias -- high variance in detection accuracy across target groups, shows that data distribution in these unbalanced datasets is a critical factor. Hate speech detection on a target group is more challenging with fewer corresponding training data points. Additionally, stylistic differences in hateful and offensive text against different minorities also plays a role in poor performance, as discussed in Section \ref{sec2}.

We propose to address this using a token level hate sense disambiguation based approach. Benign tokens like \textit{woman} and \textit{gay} can be hateful when targeting a particular group and used in malicious context. To distinguish the hateful application from benign, we implement a token-level model which predicts the hate sense (distribution over class labels) at every time-step while predicting the overall hate speech class. Subsequently, the classifier considers hate sense augmented token representations, allowing a more general detection solution. Experimentally, we show that our approach leads to a more balanced performance with a 30\% drop in variance across target groups and has an at least 4\% greater average-across-target groups performance than a BERT-based baseline.

In summary, the contribution of this paper include:
\newline
(1) We are the first to highlight a crucial problem in NLP models having an unbalanced hate speech detection capabilities across different target groups. \newline
(2) We propose a zero-shot token level hate sense disambiguation technique to address this. \newline
(3) Our technique leads to an absolute 3\% improvement in average target detection accuracy with a significant drop in group-wise performance variance. 
\section{Motivation and Analysis}
\label{sec2}
In this section, we study the performance of a BERT model trained on \citet{vidgen2020learning}. 
\newline
\textbf{Biased Performance} The BERT hate speech detection model has a biased performance as seen in Table \ref{tab:analysis}. For instance, model accuracy on the Gay Men target group is 43\% which is almost half of 85\% on South Asian's. We hypothesize that these results are due to two factors: (1) Training data available for each target group; (2) Stylistic differences in hate text used across target groups. 

\begin{table}[t]
\centering
\resizebox{\columnwidth}{!}{%
\begin{tabular}{lcc|c}
\toprule
\textbf{Target Group}    & \textbf{Training Data}   & \textbf{Word Overlap}  & \textbf{Eval Accuracy}   \\ 
\midrule
Women            & 1652           & 0.65      &      0.73             \\
Black             & 1580            &   0.79     &       0.81     \\
Jew          & 891            &    0.70    &        0.83          \\
Muslim         & 779            &    0.66  &       0.79           \\
Transgender         & 640            &  0.64    &      0.75    \\
Gay          & 580            &   0.71    &     0.67           \\
Immigrants   & 545            &   0.58    &         0.66       \\
Refugee              & 376            &   0.57     &    0.77\\
Disable          & 374            &    0.58    &    0.83    \\
South Asian           & 274            &   0.51     &   0.86             \\
Arab          & 262            &     0.53   &       0.82    \\
Gay Men   & 217            &      0.43  &       0.43        \\
Black Women              &     144        &     0.45  &     0.41            \\
East Asian           &       144      &    0.47    &    0.74      \\
Hispanic           &       57      &    0.15    &   0.60      \\
\bottomrule             
\end{tabular}%
}
\caption{Performance of a BERT model on different target groups in \citet{vidgen2020learning}. Statistics on specific number of training data points and fraction of word overlap are also provided.}
\label{tab:analysis}
\end{table}

\begin{figure}[!t]
\includegraphics[width=1.0\columnwidth]{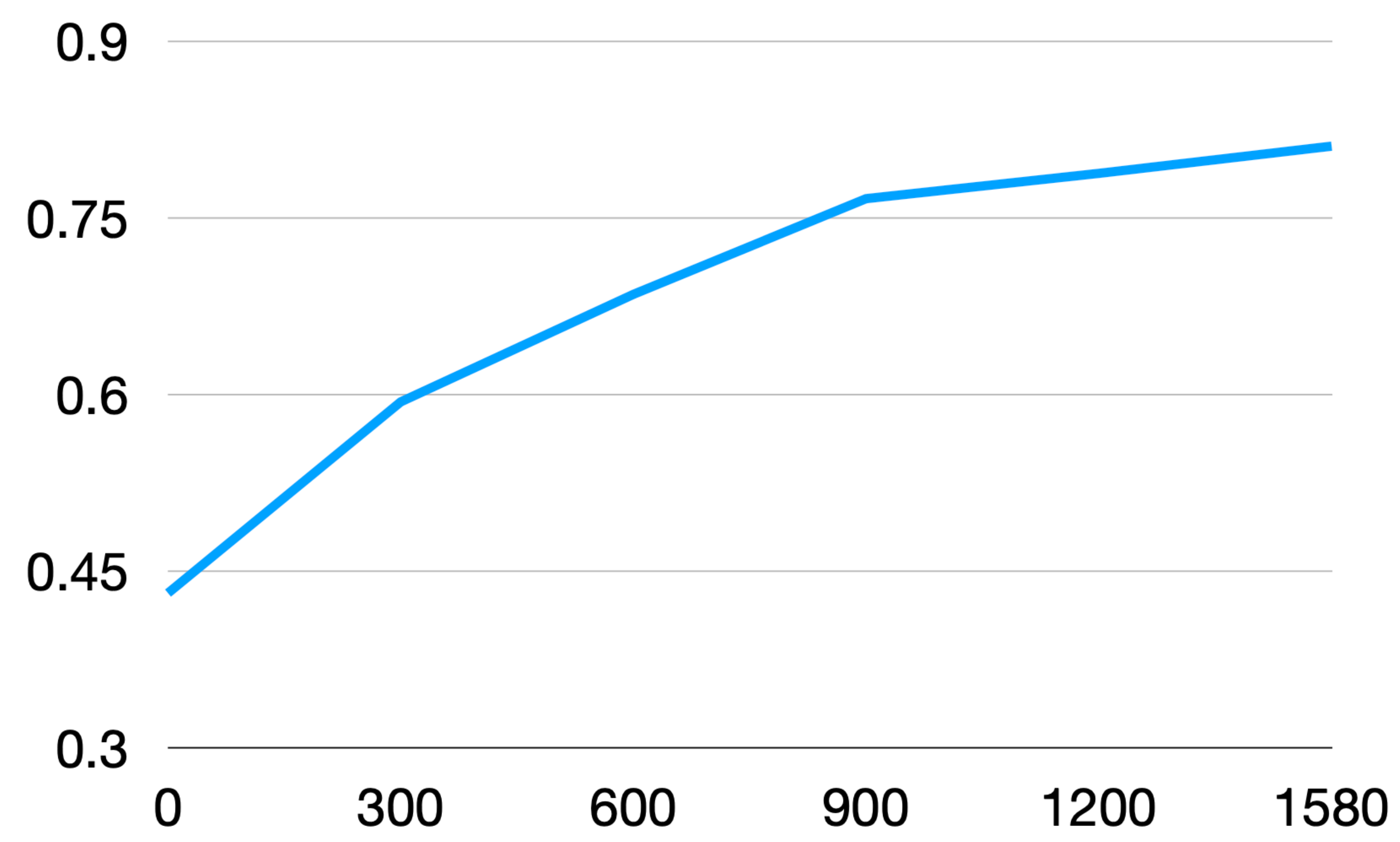}
\caption{Model performance on the "Black" target group with an increasing number of "Black" training data points.}
\label{fig:training_data}
\end{figure}
\textbf{Training Data}
We investigate the impact of training data available for every target group and the corresponding model performance. In particular, we look at the model performance on the Black target group with an increasing number of corresponding training data available. Figure \ref{fig:training_data} shows that performance on the Black target group improves with an increase in training data.

On the complete dataset we also see that target groups with more training data such as Black, Jew and Muslim target groups (Table \ref{tab:analysis}) have a higher test performance than Gay Men, Black Women and Hispanics target groups which have fewer training data points. However, the size of training data is not the only deciding factor for performance. For instance, the performance on South Asian and Arab target groups is much higher than performance against Immigrant and Women target groups, the latter with far more training data. Overall, training data is an important but not exclusive factor in hate speech detection performance across target groups.

\textbf{Stylistic Differences}
Hateful text varies according to the intended target group, hence making such datasets a mixture of unique sub-domains. Such stylistic differences have the potential to cause a variance in performance across target groups. Table \ref{tab:analysis} reports the token overlap for the most frequent tokens used against different target groups with most frequent tokens used in the rest of the data. A higher word overlap for Black, Jew, Women and Muslim target groups corresponds to a high test accuracy, while a lower word overlap for Immigrant, Hispanics and Black Women target groups corresponds to a lower test performance. Performance on Arabs and South Asians target groups with low word overlap is higher than the performance against Gay target group which has a higher word overlap. Overall, the stylistic differences does not explain all the bias but is a strong factor.

\begin{figure*}[!h]
\centering
\includegraphics[width=0.95\textwidth]{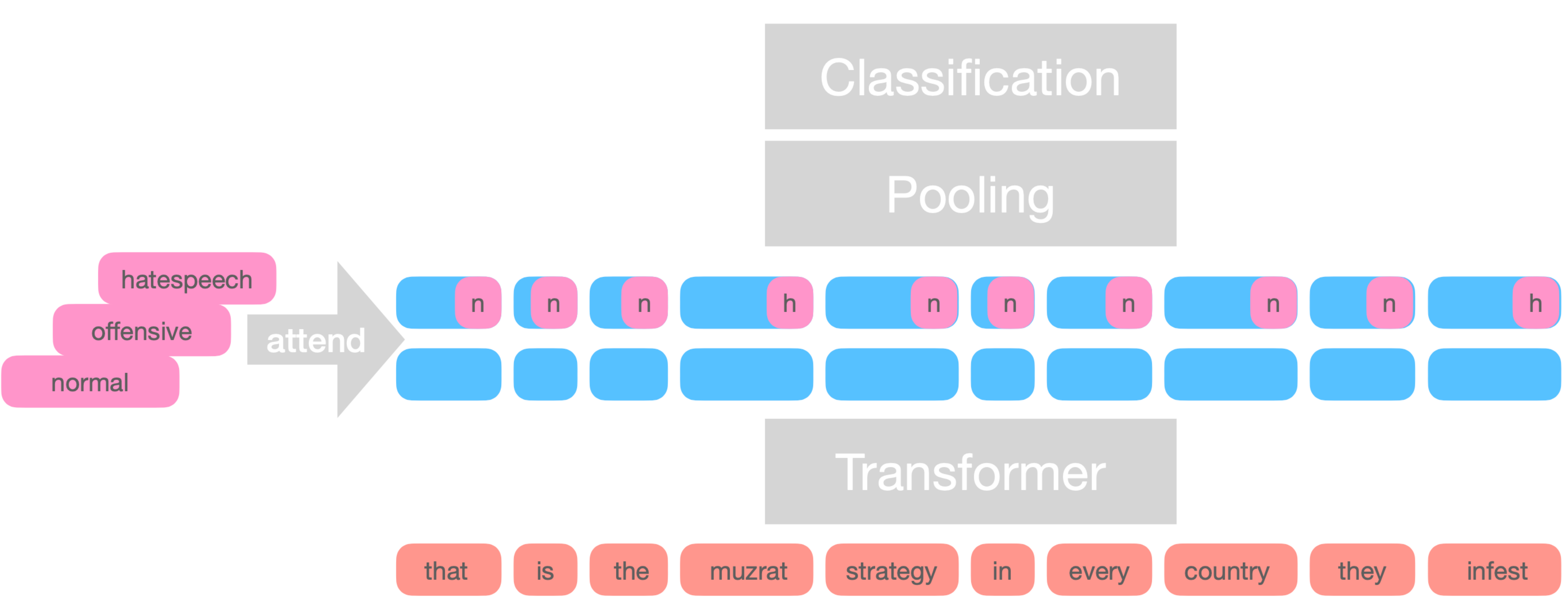}
\caption{Figure illustrating the flow of our model.}
\label{fig:model_arch}
\end{figure*}

\section{Towards Unbiased Modeling}

In this section, we propose a token-level model which performs sense disambiguation enroute to the overall hate speech prediction. Specifically, we develop our model to detect hate speech related senses for all tokens using their contextual information. Apart from augmenting the tokens' hidden representations with their sense representations, our model is regularized to force the consensus of the token level hate sense predictions to agree with the target hate sense. This enables our model to rely on general signals for the overall hate speech detection compared to vanilla models.\newline 
\textbf{Architecture} Figure \ref{fig:model_arch} shows our model architecture. We consider a Transformer based text encoder $E$ to represent our inputs. For a potential hateful text input $x = \{x_1, x_2, ..., x_n\}$, our model produces representations for every token $E(x) = [E(x_1), E(x_2), ..., E(x_n)]$. The named hate speech classes $C = \{c_1, c_2, ..., c_k\}$ are also represented by using $E$ on their corresponding class names to get $\{E(c_1), E(c_2), ..., E(c_k)\}$ through encoding and subsequent pooling. 

Every hidden state $E(x_i)$ in $E(x)$ is attended to the class representations $\{E(c_1), E(c_2), ..., E(c_k)\}$. The hate sense $s_i$ for the hidden state $E(x_i)$ is categorized as token $x_i$'s sense, where :
\begin{equation}
s_i = \operatorname*{arg\,max}_j \frac{\exp(cos(E(x_i),E(c_j)))}{\sum_{l=1}^{k}\exp(cos(E(x_i),E(c_l)))}
\end{equation}

The final prediction, $f(x) = C(E(x))$ with $C$ a Multi-layer Perceptron and Pooling classifier and $E$ the encoder utilize this sense prediction $s$ and attended hidden representations. Specifically, $f(x) = C([E(x_1) + E(c_{s_{1}}), E(x_2) + E(c_{s_{2}}), ..., E(x_n)+ E(c_{s_{n}})])$ where a max-pooling and multi-layer perceptron classifier $C$ is applied to the attended representations. 
\newline
\textbf{Optimization} In addition to minimizing the final loss $L(f(x),y)$, we enforce constraints on the token level sense predictions having their consensus match the final hate speech label ($M$ selects the max occurring hateful sense):
\begin{equation}
    L(M(s_1, s_2, ..., s_n),y)
\end{equation}
We enforce the number of unique hateful senses to be minimized ($U$ selects all unique hateful senses):
\begin{equation}
    ||U(s_1, s_2, ..., s_n)||_{L_1}
\end{equation}

We hypothesize that our sense prediction approach, implemented through these constraints, models sentence semantics better to allow for a robust hate speech detection.

\section{Experiments}

We report performance across target groups on two public datasets  Learning from the Worst \emph{LearningWorst} \citep{vidgen2020learning} and \emph{HateXplain} \citep{mathew2020hatexplain}. Both these datasets have annotations on the target groups.\footnote{To the best of our knowledge the performance per target group has not been previously reported.} Tables \ref{tab:results1} and \ref{tab:results2} list the target groups in the respective datasets.\footnote{We consider all target groups with at least 25 data points in the test set.}

We consider a BERT document level text-classification model \citep{devlin-etal-2019-bert} for hate speech detection. We develop our token-level sense disambiguation model on top of this model. The models are implemented using the Huggingface library \citep{wolf2019huggingface}.

\begin{table}[h]
\centering
\resizebox{\columnwidth}{!}{%
\begin{tabular}{lcc}
\toprule
\textbf{Target Group}    & \textbf{Baseline Performance}   & \textbf{Method Performance}\\ 
\midrule
Women            & \textbf{0.73}          & 0.71                 \\
Black             & 0.81            &   \textbf{0.83}        \\
Jew          & 0.83            &    \textbf{0.85}             \\
Muslim         & 0.79            &    \textbf{0.82}           \\
Transgender         & 0.75            &  \textbf{0.78}      \\
Gay          & 0.67           &   \textbf{0.74}             \\
Immigrants   & 0.66            &   \textbf{0.69}          \\
Refugee              & 0.77            &   0.77     \\
Disable          & \textbf{0.83}            &    0.80       \\
South Asian           & \textbf{0.86}            &   0.82                 \\
Arab          & 0.82            &     \textbf{0.85}       \\
Gay Men   & 0.43            &      \textbf{0.57}        \\
Black Women              &     0.41        &     \textbf{0.59}           \\
East Asian           &       0.74      &    \textbf{0.79}         \\
Hispanic           &       \textbf{0.60}      &    0.56   \\
\midrule
Test Performance & \textbf{0.78} & 0.77 \\
Average (across targets) & 0.71 & \textbf{0.74} \\
Performance Variance & 0.14 & \textbf{0.10} \\
\bottomrule             
\end{tabular}%
}
\caption{Comparison of baseline BERT and token-level classification model on \emph{LearningWorst}.}
\label{tab:results1}
\end{table}

\begin{table}[h]
\centering
\resizebox{\columnwidth}{!}{%
\begin{tabular}{lcc}
\toprule
\textbf{Target Group}    & \textbf{Baseline Performance}   & \textbf{Method Performance}\\ 
\midrule
African          & 0.54            &     \textbf{0.75}      \\
Jewish          & 0.57            &     \textbf{0.79}     \\
Islam   & \textbf{0.75}           &      0.71         \\
Homosexual              &     \textbf{0.76}        &    0.73            \\
Women           &       \textbf{0.63}      &    0.61         \\
Arab           &       0.71      &    \textbf{0.74}   \\
\midrule
Test Performance & \textbf{0.77} & 0.76 \\
Average (across targets) & 0.66 & \textbf{0.72} \\
Performance Variance & 0.09 & \textbf{0.06} \\
\bottomrule             
\end{tabular}%
}
\caption{Comparison of baseline BERT and token-level classification model on \emph{HateXplain}.}
\label{tab:results2}
\end{table}

\paragraph{Results} Tables \ref{tab:results1} and \ref{tab:results2} report the results of the baseline BERT method and our debiasing approach on the \emph{LearningWorst} and \emph{HateXplain} datasets respectively. 
Table \ref{tab:results1} reports how our method is able to reduce the variance in accuracy across different target groups while improving the average accuracy across all target groups. The performance on several poor performing target groups like Gay Men, Black Women and Immigrants is significantly improved by our method. The lowest accuracy is now on the  Hispanics target group at 56\% which is significantly higher than the original lowest of 41\% on the Black Women target group.  This balanced performance comes at a slight cost of 1\% drop in the overall test accuracy.
Similarly, Table \ref{tab:results2} reports how our method is able to reduce the variance in accuracy across different target groups while improving the average performance across all target groups. The accuracy on poor performing African and Jewish target groups is significantly improved by our method. The lowest performance is now on Women target group at 60\% which is higher than the previous lowest on African target group at 54\%. The balanced performance comes at a slight cost of 1\% drop in the overall test accuracy. 

Our method is effective in reducing the bias by performing better in scenarios with fewer training data points and greater stylistic differences.
\section{Related Work}
\textbf{Bias in hate speech Detection} The growth of hate and abuse online has inspired the collection of several datasets to study the phenomenon  \citep{waseem2016hateful,waseem2016you,davidson2017automated,founta2018large,mandl2019overview,mandl2020overview,kumar2018benchmarking,zampieri-etal-2019-predicting,mathew2020hatexplain, toutanova2021proceedings}.  While these datasets form numerous benchmarks to compare machine learning solutions, several issues have been identified with
hate speech training datasets -- lack of linguistic variety and annotations \citep{vidgen2019challenges,poletto2021resources}. In particular sampling data by searching for keywords can lead to the collection of a biased dataset \citep{vidgen2019much,wiegand2019detection}. In our work we are identifying a bias in peformance across target groups for models trained on hate speech datasets. This work falls in a broader category on fairness and debiasing across various other language tasks \citep{sun2019mitigating,chang2019bias,clark2019don,schuster2019towards}.
\newline
\textbf{Few Shot Sense Detection} Word Sense Detection \citep{miller1993semantic} is a long standing task of identifying the meaning of a word in a specific text. Recent methods \citep{huang2019glossbert,blevins2020moving,bevilacqua2020breaking} have outperformed human performance on sense detection \citep{navigli2009word}. In scenarios where certain senses are rare, the performance of typical models is not optimal and a BERT based description of the senses helps alleviate the low resource problem \citep{blevins2021fews}. In this work, we focus on identifying hateful senses as annotated in the training datasets, using their BERT representations. Despite having no sense annotations, we use the class names to assign token level senses. 

\section{Discussion}
This paper demonstrates that models trained on hate speech datasets may have biased performance across different target groups. Our analysis shows that additional training data related to a target group is beneficial, highlighting the need for a more balanced collection of hateful text. We suggest a sense-based solution to address this issue, leading to a better average performance across different target groups. 


\bibliography{anthology,custom}

\begin{thebibliography}{27}
\expandafter\ifx\csname natexlab\endcsname\relax\def\natexlab#1{#1}\fi

\bibitem[{Bevilacqua and Navigli(2020)}]{bevilacqua2020breaking}
Michele Bevilacqua and Roberto Navigli. 2020.
\newblock Breaking through the 80\% glass ceiling: Raising the state of the art
  in word sense disambiguation by incorporating knowledge graph information.
\newblock In \emph{Proceedings of the 58th Annual Meeting of the Association
  for Computational Linguistics}, pages 2854--2864.

\bibitem[{Blevins et~al.(2021)Blevins, Joshi, and
  Zettlemoyer}]{blevins2021fews}
Terra Blevins, Mandar Joshi, and Luke Zettlemoyer. 2021.
\newblock Fews: Large-scale, low-shot word sense disambiguation with the
  dictionary.
\newblock \emph{arXiv preprint arXiv:2102.07983}.

\bibitem[{Blevins and Zettlemoyer(2020)}]{blevins2020moving}
Terra Blevins and Luke Zettlemoyer. 2020.
\newblock Moving down the long tail of word sense disambiguation with
  gloss-informed biencoders.
\newblock \emph{arXiv preprint arXiv:2005.02590}.

\bibitem[{Chang et~al.(2019)Chang, Prabhakaran, and Ordonez}]{chang2019bias}
Kai-Wei Chang, Vinod Prabhakaran, and Vicente Ordonez. 2019.
\newblock Bias and fairness in natural language processing.
\newblock In \emph{Proceedings of the 2019 Conference on Empirical Methods in
  Natural Language Processing and the 9th International Joint Conference on
  Natural Language Processing (EMNLP-IJCNLP): Tutorial Abstracts}.

\bibitem[{Clark et~al.(2019)Clark, Yatskar, and Zettlemoyer}]{clark2019don}
Christopher Clark, Mark Yatskar, and Luke Zettlemoyer. 2019.
\newblock Don't take the easy way out: Ensemble based methods for avoiding
  known dataset biases.
\newblock \emph{arXiv preprint arXiv:1909.03683}.

\bibitem[{Davidson et~al.(2017)Davidson, Warmsley, Macy, and
  Weber}]{davidson2017automated}
Thomas Davidson, Dana Warmsley, Michael Macy, and Ingmar Weber. 2017.
\newblock Automated hate speech detection and the problem of offensive
  language.
\newblock In \emph{Proceedings of the International AAAI Conference on Web and
  Social Media}, volume~11.

\bibitem[{Devlin et~al.(2019)Devlin, Chang, Lee, and
  Toutanova}]{devlin-etal-2019-bert}
Jacob Devlin, Ming-Wei Chang, Kenton Lee, and Kristina Toutanova. 2019.
\newblock \href {https://doi.org/10.18653/v1/N19-1423} {{BERT}: Pre-training of
  deep bidirectional transformers for language understanding}.
\newblock In \emph{Proceedings of the 2019 Conference of the North {A}merican
  Chapter of the Association for Computational Linguistics: Human Language
  Technologies, Volume 1 (Long and Short Papers)}, pages 4171--4186,
  Minneapolis, Minnesota. Association for Computational Linguistics.

\bibitem[{Founta et~al.(2018)Founta, Djouvas, Chatzakou, Leontiadis, Blackburn,
  Stringhini, Vakali, Sirivianos, and Kourtellis}]{founta2018large}
Antigoni~Maria Founta, Constantinos Djouvas, Despoina Chatzakou, Ilias
  Leontiadis, Jeremy Blackburn, Gianluca Stringhini, Athena Vakali, Michael
  Sirivianos, and Nicolas Kourtellis. 2018.
\newblock Large scale crowdsourcing and characterization of twitter abusive
  behavior.
\newblock In \emph{Twelfth International AAAI Conference on Web and Social
  Media}.

\bibitem[{Huang et~al.(2019)Huang, Sun, Qiu, and Huang}]{huang2019glossbert}
Luyao Huang, Chi Sun, Xipeng Qiu, and Xuanjing Huang. 2019.
\newblock Glossbert: Bert for word sense disambiguation with gloss knowledge.
\newblock \emph{arXiv preprint arXiv:1908.07245}.

\bibitem[{Kumar et~al.(2018)Kumar, Ojha, Malmasi, and
  Zampieri}]{kumar2018benchmarking}
Ritesh Kumar, Atul~Kr Ojha, Shervin Malmasi, and Marcos Zampieri. 2018.
\newblock Benchmarking aggression identification in social media.
\newblock In \emph{Proceedings of the First Workshop on Trolling, Aggression
  and Cyberbullying (TRAC-2018)}, pages 1--11.

\bibitem[{Mandl et~al.(2020)Mandl, Modha, Kumar~M, and
  Chakravarthi}]{mandl2020overview}
Thomas Mandl, Sandip Modha, Anand Kumar~M, and Bharathi~Raja Chakravarthi.
  2020.
\newblock Overview of the hasoc track at fire 2020: Hate speech and offensive
  language identification in tamil, malayalam, hindi, english and german.
\newblock In \emph{Forum for Information Retrieval Evaluation}, pages 29--32.

\bibitem[{Mandl et~al.(2019)Mandl, Modha, Majumder, Patel, Dave, Mandlia, and
  Patel}]{mandl2019overview}
Thomas Mandl, Sandip Modha, Prasenjit Majumder, Daksh Patel, Mohana Dave,
  Chintak Mandlia, and Aditya Patel. 2019.
\newblock Overview of the hasoc track at fire 2019: Hate speech and offensive
  content identification in indo-european languages.
\newblock In \emph{Proceedings of the 11th forum for information retrieval
  evaluation}, pages 14--17.

\bibitem[{Mathew et~al.(2020)Mathew, Saha, Yimam, Biemann, Goyal, and
  Mukherjee}]{mathew2020hatexplain}
Binny Mathew, Punyajoy Saha, Seid~Muhie Yimam, Chris Biemann, Pawan Goyal, and
  Animesh Mukherjee. 2020.
\newblock Hatexplain: A benchmark dataset for explainable hate speech
  detection.
\newblock \emph{arXiv preprint arXiv:2012.10289}.

\bibitem[{Miller et~al.(1993)Miller, Leacock, Tengi, and
  Bunker}]{miller1993semantic}
George~A Miller, Claudia Leacock, Randee Tengi, and Ross~T Bunker. 1993.
\newblock A semantic concordance.
\newblock In \emph{Human Language Technology: Proceedings of a Workshop Held at
  Plainsboro, New Jersey, March 21-24, 1993}.

\bibitem[{Navigli(2009)}]{navigli2009word}
Roberto Navigli. 2009.
\newblock Word sense disambiguation: A survey.
\newblock \emph{ACM computing surveys (CSUR)}, 41(2):1--69.

\bibitem[{Poletto et~al.(2021)Poletto, Basile, Sanguinetti, Bosco, and
  Patti}]{poletto2021resources}
Fabio Poletto, Valerio Basile, Manuela Sanguinetti, Cristina Bosco, and Viviana
  Patti. 2021.
\newblock Resources and benchmark corpora for hate speech detection: a
  systematic review.
\newblock \emph{Language Resources and Evaluation}, 55(2):477--523.

\bibitem[{Schuster et~al.(2019)Schuster, Shah, Yeo, Filizzola, Santus, and
  Barzilay}]{schuster2019towards}
Tal Schuster, Darsh~J Shah, Yun Jie~Serene Yeo, Daniel Filizzola, Enrico
  Santus, and Regina Barzilay. 2019.
\newblock Towards debiasing fact verification models.
\newblock \emph{arXiv preprint arXiv:1908.05267}.

\bibitem[{Sun et~al.(2019)Sun, Gaut, Tang, Huang, ElSherief, Zhao, Mirza,
  Belding, Chang, and Wang}]{sun2019mitigating}
Tony Sun, Andrew Gaut, Shirlyn Tang, Yuxin Huang, Mai ElSherief, Jieyu Zhao,
  Diba Mirza, Elizabeth Belding, Kai-Wei Chang, and William~Yang Wang. 2019.
\newblock Mitigating gender bias in natural language processing: Literature
  review.
\newblock \emph{arXiv preprint arXiv:1906.08976}.

\bibitem[{Toutanova et~al.(2021)Toutanova, Rumshisky, Zettlemoyer, Hakkani-Tur,
  Beltagy, Bethard, Cotterell, Chakraborty, and
  Zhou}]{toutanova2021proceedings}
Kristina Toutanova, Anna Rumshisky, Luke Zettlemoyer, Dilek Hakkani-Tur,
  Iz~Beltagy, Steven Bethard, Ryan Cotterell, Tanmoy Chakraborty, and Yichao
  Zhou. 2021.
\newblock Proceedings of the 2021 conference of the north american chapter of
  the association for computational linguistics: Human language technologies.
\newblock In \emph{Proceedings of the 2021 Conference of the North American
  Chapter of the Association for Computational Linguistics: Human Language
  Technologies}.

\bibitem[{Vidgen et~al.(2019{\natexlab{a}})Vidgen, Harris, Nguyen, Tromble,
  Hale, and Margetts}]{vidgen2019challenges}
Bertie Vidgen, Alex Harris, Dong Nguyen, Rebekah Tromble, Scott Hale, and Helen
  Margetts. 2019{\natexlab{a}}.
\newblock Challenges and frontiers in abusive content detection.
\newblock In \emph{Proceedings of the third workshop on abusive language
  online}, pages 80--93.

\bibitem[{Vidgen et~al.(2019{\natexlab{b}})Vidgen, Margetts, and
  Harris}]{vidgen2019much}
Bertie Vidgen, Helen Margetts, and Alex Harris. 2019{\natexlab{b}}.
\newblock How much online abuse is there.
\newblock \emph{Alan Turing Institute}.

\bibitem[{Vidgen et~al.(2020)Vidgen, Thrush, Waseem, and
  Kiela}]{vidgen2020learning}
Bertie Vidgen, Tristan Thrush, Zeerak Waseem, and Douwe Kiela. 2020.
\newblock Learning from the worst: Dynamically generated datasets to improve
  online hate detection.
\newblock \emph{arXiv preprint arXiv:2012.15761}.

\bibitem[{Waseem(2016)}]{waseem2016you}
Zeerak Waseem. 2016.
\newblock Are you a racist or am i seeing things? annotator influence on hate
  speech detection on twitter.
\newblock In \emph{Proceedings of the first workshop on NLP and computational
  social science}, pages 138--142.

\bibitem[{Waseem and Hovy(2016)}]{waseem2016hateful}
Zeerak Waseem and Dirk Hovy. 2016.
\newblock Hateful symbols or hateful people? predictive features for hate
  speech detection on twitter.
\newblock In \emph{Proceedings of the NAACL student research workshop}, pages
  88--93.

\bibitem[{Wiegand et~al.(2019)Wiegand, Ruppenhofer, and
  Kleinbauer}]{wiegand2019detection}
Michael Wiegand, Josef Ruppenhofer, and Thomas Kleinbauer. 2019.
\newblock Detection of abusive language: the problem of biased datasets.
\newblock In \emph{Proceedings of the 2019 conference of the North American
  Chapter of the Association for Computational Linguistics: human language
  technologies, volume 1 (long and short papers)}, pages 602--608.

\bibitem[{Wolf et~al.(2019)Wolf, Debut, Sanh, Chaumond, Delangue, Moi, Cistac,
  Rault, Louf, Funtowicz et~al.}]{wolf2019huggingface}
Thomas Wolf, Lysandre Debut, Victor Sanh, Julien Chaumond, Clement Delangue,
  Anthony Moi, Pierric Cistac, Tim Rault, R{\'e}mi Louf, Morgan Funtowicz,
  et~al. 2019.
\newblock Huggingface's transformers: State-of-the-art natural language
  processing.
\newblock \emph{arXiv preprint arXiv:1910.03771}.

\bibitem[{Zampieri et~al.(2019)Zampieri, Malmasi, Nakov, Rosenthal, Farra, and
  Kumar}]{zampieri-etal-2019-predicting}
Marcos Zampieri, Shervin Malmasi, Preslav Nakov, Sara Rosenthal, Noura Farra,
  and Ritesh Kumar. 2019.
\newblock \href {https://doi.org/10.18653/v1/N19-1144} {Predicting the type and
  target of offensive posts in social media}.
\newblock In \emph{Proceedings of the 2019 Conference of the North {A}merican
  Chapter of the Association for Computational Linguistics: Human Language
  Technologies, Volume 1 (Long and Short Papers)}, pages 1415--1420,
  Minneapolis, Minnesota. Association for Computational Linguistics.

\end{thebibliography}
\bibliographystyle{acl_natbib}




\end{document}